# Design of Arabic Diacritical Marks

Mohamed Hssini[1] and Azzeddine Lazrek[2]

[1] Department of Computer Science, Faculty of Sciences, University Cadi Ayyad
Marrakech, Morocco
m.hssini@ucam.ac.ma

[2] Department of Computer Science, Faculty of Sciences, University Cadi Ayyad
Marrakech, Morocco
lazrek@ucam.ac.ma

**Abstract**
Diacritical marks play a crucial role in meeting the criteria of usability of typographic text, such as: homogeneity, clarity and legibility. To change the diacritic of a letter in a word could completely change its semantic. The situation is very complicated with multilingual text. Indeed, the problem of design becomes more difficult by the presence of diacritics that come from various scripts; they are used for different purposes, and are controlled by various typographic rules. It is quite challenging to adapt rules from one script to another. This paper aims to study the placement and sizing of diacritical marks in Arabic script, with a comparison with the Latin's case. The Arabic script is cursive and runs from right-to-left; its criteria and rules are quite distinct from those of the Latin script. In the beginning, we compare the difficulty of processing diacritics in both scripts. After, we will study the limits of Latin resolution strategies when applied to Arabic. At the end, we propose an approach to resolve the problem for positioning and resizing diacritics. This strategy includes creating an Arabic font, designed in OpenType format, along with suitable justification in $T_EX$.

***Keywords***: *Arabic calligraphy, Diacritical mark, Justification, Multilevel ligature, Multilingual, OpenType.*

## 1. Introduction

The typographical choices can make or break the success of a digital document. If the text is difficult to read or does not look satisfactory, users will question the validity of its content [1] or simply move to another document that is more user-friendly. Digital typography, as an art, has its elements, its principles and attributes [2]; controlled by rules, but also limited by constraints. It's, as a technique, based on the concept of digital fonts. A font is a set of graphical presentations of characters, called *glyphs*, with some controlling rules. The *rendering engine* gathers them to display the words and lines that make up the text. Some of the constraints facing the typography are technical in nature: the material resources are limited enough to satisfy an aesthetic need. In multilingual digital typesetting, the principles and attributes of design are risky because of the conflicting rules and mechanisms that control and affect each script. Diacritical marks are not an exception and do not escape this rule. For example, the meaning of diacritics varies considerably according to the language. A diacritic is a sign accompanying a group of letters or one letter, as the circumflex accent "^" on the "a" producing "â". Diacritics are often placed above the letter, but they can also be placed below, in, through, before, after or around a glyph.

Diacritics have common roles between the different languages of the world, such as:
- to provide a reading;
- to amend the phonetic value of a letter;
- to avoid ambiguity between two homographs.

However, the Arabic diacritics have an additional role, which is to fill space. This is a task influenced by different effects such as: multilevel and justification of Arabic text. These contextual varieties that control the choice of Arabic diacritics sizes have been simplified in Arabic printed model.

The problem, studied in this paper, is how to establish a mechanism for extending the Arabic diacritics to adapt the calligraphic design of marks to the technical constraints of font formats: a question that has not been discussed before.

The proposed solution is based on the determination of diacritical size, on neighborhood context consisting of the base letter and the next letter that follow it in the same word.

To address it in this paper, we will discuss the following topics: first, we compare the origins, roles, and the Unicode encoding that governs the computing treatment of diacritics. Second, we compare design problems of diacritics in the Arabic script which have arisen for Latin script. Third, we identify strategies offered by OpenType to solve this problem and examine their capability in Arabic. Fourth, we consider our proposed font and algorithm as a way to solve the positioning and resizing of diacritics, in an Arabic font developed in OpenType format. We end with some conclusions and perspectives.

## 2. Diacritical Marks

### 2.1 History

There are some similarities between the history of Arabic diacritical marks and the history of the Latin's one. However, the differences are many and varied, reflecting not only the linguistic and graphic features of each script, but differences between the principles on which are based the two civilizations to which they belong. So, we find that the first Latin diacritic appeared among the ancient Greeks and Romans. They were developed and distributed in various European scripts. The diacritical marks generally descend from letters that were placed above another letter. The addition of diacritics was a choice among four options to overcome the shortcomings of a language belonging to the Latin script [3]. The others were: to add another letter, to combine two or more letters, or to use the apostrophe. The addition of diacritics in Latin script evolved over time [4]. In periods of colonization, Latin diacritics have been used to expand the Latin alphabet for writing non-Roman languages. When a language has more fundamentally different sounds – phonemes – than base letters, it can invent new letters or adopt letters from other alphabets. The solution that is so much more common is to add diacritics on the letters, often imitating the spellings of other languages [5].

When the holy Quran was documented, the Arabic alphabet had neither dots nor diacritics. Both of them were added successively during later periods. In Arabic writing, the same base glyph can represent multiple letters and the same word without vowels can represent multiple semantics [6]. The reading difficulties caused by confusion between consonants of the same shape and between words of same shape, the lack of scoring short vowels led to the invention of diacritical signs to become fixed and facilitate reading. At first, short vowels were added by placing color dots above or below letters. This usage changed and led to the current practice of marking vowels by small signs. Their shapes origins are from corresponding long vowels letters. Letters represented by same base glyph are differentiated by adding a number of dots above or below glyph.

After, some diacritics are added to the Arabic base alphabet to form new letters used to write some languages as: Old Turkic, Urdu, and Farsi [7].

2.2 Classification

There are three kinds of Arabic diacritics, according to their typography [8] (see Fig. 1 to 8):

- *Language's diacritics*: differentiate the letter's consonants, are very important for semantic. They appear as:
  o **Diacritics above**: placed above a letter, as Fatha, Damma, Soukoun, or Shadda.

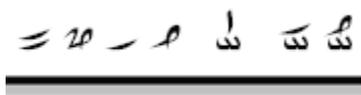

Fig. 1.    Arabic diacritics above

  o **Diacritics below**: placed below a letter, as Kasra or Kasratan.

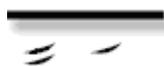

Fig. 2.    Arabic diacritics below

  o **Diacritics through**: placed through a letter, as Wasl.

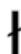

Fig. 3.    Jarrat Wasl through Alef

- *Aesthetic diacritics*: often filled space created when extending some letters, to improve the aesthetic.

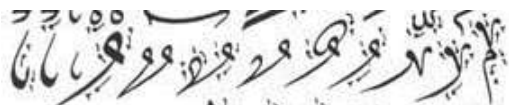

Fig. 4.    Arabic aesthetic diacritics

- *Explanatory diacritics*: positioned to distinguish the Muhmal and Muajam letters. Arabic letters are divided into two categories: Muhmal letters without dots and Muajam letters, based on Muhmal ones, but containing dots.

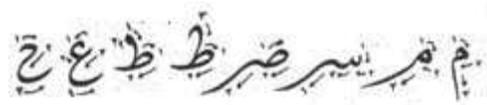

Fig. 5.    Arabic explanatory diacritics

The features of Latin diacritics affect their positions, and can be presented according to their placements on their base letters, as follow:

- *Superscript-diacritics*:

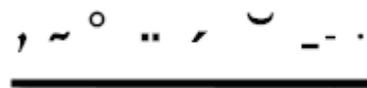

Fig. 6.    Latin diacritics above

- *Subscript-diacritics*:

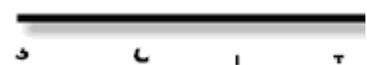

Fig. 7.    Latin diacritics below

- *Others diacritics*: there are other diacritics that are positioned through, before, after, or around a letter's glyph.

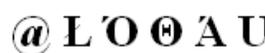

Fig. 8.    Others Latin diacritics

Latin diacritics can also be classified according to their design or their Unicode encoding [4].

## 3. Diacritics in Unicode

Before Unicode, there were limits the number of characters that could be encoded. The set of standard ASCII characters is 128 characters, 95 printable characters, including 52 alphabetic characters (the 26 Latin letters in uppercase and lowercase), but no accented letters. There are several other character sets, called *ASCII extended*, which include 256 characters, with the additional 128 characters used to represent particular vowels and consonants of the Latin alphabet with diacritics or occurring in other alphabets [9].

3.1 Encoding

Unicode is a character encoding standard that defines a consistent way of encoding multilingual texts and facilitates the exchange of textual data. It could, in theory, encode all characters used by all the written languages of the world (more than one million characters are reserved for this purpose). All characters, regardless of the script in which they are used, are accessible without escape sequences. The Unicode encoding treats alphabetic characters, ideographic characters and symbols in an equivalent manner, with the result that they can coexist in any order with equal ease. For each character, the standard Unicode allocate a unique numeric value and attribute a single block name. As such, it differs little from other standards. However, Unicode specify other information pivotal to guarantee that the encoded text will be clear to read: the case of the characters (if they have case), their properties, and directionality. Unicode also defines semantic information and includes correspondence tables of conversions between Unicode and other important character sets. In Unicode, diacritics appear as a category of combinatorial signs [10].

### 3.2 Arabic diacritics in Unicode

In Unicode, Arabic diacritics are treated in two different ways:
- Diacritics encoded in conjuncture with his basic letter, such as: alef with madda.
- Diacritics encoded alone.

The encoding of Arabic diacritic is distributed in the four following blocks [11] [12]:
- Arabic (0600–06FF): this range includes the standard letters and diacritics.
- Arabic Supplement (0750–077F): this range incorporates Arabic diacritics in conjuncture with their basic letters used for extending Arabic to writing others scripts.
- Arabic Presentation Forms-A (FB50–FDFF): this range represents Arabic diacritics considered in isolation.
- Arabic Presentation Forms-B (FE70–FEFF): this range adds spacing forms of Arabic diacritics.

### 3.3 Synthesis

Diacritics are the main set of combinatorial non-spacing marks. They are treated in different manners: sometimes they are encoded with their base glyphs as "à", and sometimes they are encoded separately as Arabic standard diacritics. Software that take part Arabic diacritics in rendering must accomplishes much more amplified operations than the positioning of Latin diacritics. In Arabic case, software is supposed to analyze the base character, the combining diacritic, the neighboring base glyphs and their diacritics.

In Arabic script, dots are diacritics that play the same role as Latin diacritics. The Unicode failure to encode Arabic characters with dots, as composite characters, limits the dynamicity of dots with regard to multilevel and justification.

## 4. Design and Multilingualism

Various fundamental notions underlie the domain of design, such as balance, rhythm, etc. The principles of design clash in the case of the mixture of different styles, which may differ depending upon each script. A somewhat similar situation occurs in a monolingual Arabic text where there is a change of calligraphic styles, such as at the beginning of a title or section [7].

The size of the combining Arab varies depending on the context, and depending on how to choose the allographs. This choice reflects relations between the neighboring letters specific to each calligraphic style.

### 4.1 Calligraphic styles

In Arabic calligraphy, there are various styles of writing. The main ones are: Naskh, Riqaa, Thulut, Maghrebi, and Diwani. These styles differentiate principally by [13] [14]:
- geometric shape of the letters;
- presence, shape and number of dots;
- presence, shape and number of diacritics;
- use, shape and size of Kashida.

Each writing style has its own strict rules and context (edition, illustration, architectural decoration, etc.). This study concerns only Naskh style.

### 4.2 Contextual dependence

In most scripts, such as Latin, Hebrew and Chinese, letters are in an imaginary box that can be aligned with the letter "x" [7]. However, in the Arabic script, the heights and forms of letters vary depending on the context. In general, letters have four forms: *initial*, *medial*, *final*, and *isolated* forms [8]. In calligraphy, some forms also vary according to the neighboring letters glyphs [13] (see Fig. 9).

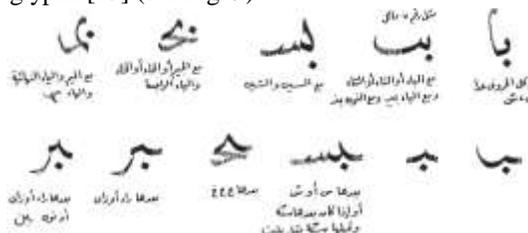

Fig. 9. Variants of Arabic letter Beh in initial and medial forms

The *spatial* properties vary between Latin and Arabic scripts. In Arabic, the definition of *bold* depends on calligraphic style. The reduction in the density of letters is created by layering or by reducing the letter's body [7]. Diacritics in the Thulut style, unlike Naskh, are designed by a pen, called a *Qalam*, with a width and slope different from those used for the body of base letters. The harmonization of a multilingual document is therefore influenced by the multitude of scripts or styles in the same language.

### 4.3 Multilevel ligature

Arabic script is cursive, letters are interrelated. In Arabic calligraphy, some letters could be combined forming *ligatures*. The contextual ligatures are needed for cursive writing. There are required during the computer processing of handwriting. There are three kinds of ligatures: *contextual*, *linguistic*, and *aesthetic*. The one unique linguistic ligature is LamAlef. An aesthetic ligature can be in two, three, or more levels, depending on the number of combined letters vertically. The aesthetic ligatures affect considerably the visibility of diacritical words, they are optional. There are chosen sometimes for justification in order to contracting the word. There are two blocks in Unicode that includes Arabic aesthetic ligatures. As their number is very large, and they can be represented in fonts without needing to be separately encoded, Unicode has decided to not add any more. If we observe their forms of representation, aesthetic ligatures are in all forms. Arabic writing is characterized by multiple baselines, used to position letters in ligatures vertically, known as multilevel or stacking of Arabic writing. Ligatures introduce the multilevel of writing. The aesthetic ligatures were a fairly limited choice to represent the multilevel of the Arabic script: the number of combinations represented in a multilevel context is large enough that one can guess their representation in a block. This property is caused also by letters: Family Jeem (Jeem, Hah, and Khah), Meem, and Yeh rajiaa (see Fig. 10, Chawki font).

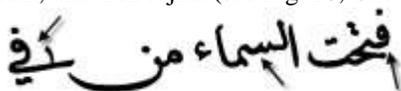

Fig. 10.    Multilevel Ligatures

Yeh rajiaa, diverted, occurs at the end of a word which is preceded by another letter which its glyph ends with the body of a Noon, or *Kaas*.

## 4.4 Kashida

Kashida is the curvilinear elongation occurred for some letters, according some situations, following some conditions, and stretching in some sizes. It is specific to the Arabic script. Unicode has included a character for Kashida (U+640 Arabic Tatweel) in order to be inserted to stretch characters. However, in calligraphy, Kashida is rather a processing to extend some letters in curvilinear form. The Kashida is characterized by the form, how depend on the writing style (see Fig. 11, By Mohamed Amzile).

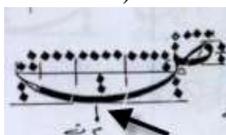

Fig. 11.    Kashida of letter Sad

**Stretching places**: Stretching by Kashida occurs in a word according to aesthetics and typographic criteria, and in respecting their roles. For example, it is a defect to superposing tow elongations in the two consecutive lines [15].

**Degree of extensibility**: The degree of extensibility of stretchable letters depends on some contextual elements [15]:

- nature of the letter to stretch;
- position of the letter in the word;
- position of the word in the line;
- level of writing that Kashida must take place;
- writing style.

**Roles**: The Kashida is used in the followed circumstances [15]:

- *Justification*: to justify Arabic text.
- *Aesthetics*: to achieve a balance and harmony between the blocks of letters in the same word.
- *legibility*: to create a void for positioning diacritics.
- *Emphasis*: related to the elongation sound of glyph extended.

## 4.5 Justification

**Justification of Latin text**:

Justification of Latin text causes the *white space* between the words and the letters, to vary, affecting the glyphs, as well as hyphenation; so that, the text fills the entire length of the line between the margins. The amount of the spacing varies between a minimal value and a maximal value when it is not possible to justify the text.

Problems related to the justification of Latin text, especially a justification of the kind made by an electronic publishing system, without correction by a human operator, are potentially quite noticeable. The most significant problems raise are: hyphenation, rivers of white, widows and orphans, and the hollow lines, which occur across blocks of text [16].

**Hyphenation**: Hyphenation permits division of a word at the end of line in order to have a better visual appearance within a text. A typographical rule requires no more than three consecutive hyphenations. Avoiding too much hyphenation in a text ensures greater fluidity of reading. There are many tools for word hyphenation, like neural networks and dictionaries, which are used to find possible hyphenation points in all words of a given language [17].

Two algorithms are used for optimizing the division of lines: Greedy algorithm and Optimum fit.

**Greedy algorithm**: This algorithm consists basically of putting as many words on a line can as possible. Then, the system repeats the same on the next line, and so on. The process is repeated until there are no more words in the paragraph. The greedy algorithm is a line-by-line process, where each line is handled individually. This algorithm is very simple and fast, and puts the list of words to be broken into a minimum number of lines. It is used by many electronic publishing systems, such as Open Office and Microsoft Word [18].

**Optimum fit algorithm**: it was employed for the first time by D. Knuth in T$_E$X. The paragraph-based algorithm uses a dynamic programming to optimize one function called the aesthetic cost function that is defined in follow. This algorithm is based on a model paragraph by an acyclic graph, where the first node is the beginning of the paragraph. In the beginning of paragraph, the algorithm creates an active node, the second node shows possible cuts, at a distance acceptable to form a line potential. This distance is defined as follows: we define *badness* from the width of inter-margins and the sum of the widths of boxes and glues component line. Each candidate line is associated with a value of *Demerits*, which is the coast in the acyclic graph where the arcs are formed with consecutive nodes [18].

**Justification of Arabic text**:

Unlike Latin justification tools, Arabic tools are:

- *Kashida*, where letters are stretched, are viewed as tools to elongate words (see Fig. 12, Chawki font).

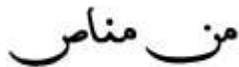
Fig. 12. Stretched glyphs

- *Ligatures*, where letters are superposed on one another, are viewed as tools to contraction words.
- *Allographs*, where one letter's glyph is substituted by another (see Fig. 13).

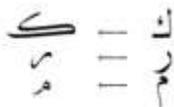
Fig. 13. Allographs of some letters

- *Moving the final letter*, in order to contract the last word in the line (see Fig. 14).

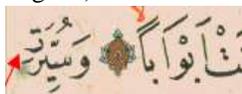
Fig. 14. Moving the final letter

- *Reduction of last letter* (see Fig. 15).

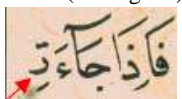
Fig. 15. Reduction of letter

- *Accumulation of words* (see Fig. 16).

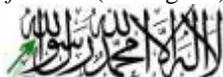
Fig. 16. Accumulation of words

- *Writing in the margin* (see Fig. 17).

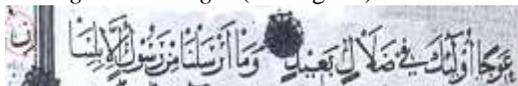
Fig. 17. Writing in margin

However, In the Arabic script, hyphenation is no longer allowed.

The optimum fit algorithm has been adapted to Arabic's needs by taking into consideration the existence of allographic variants provided by the jalt table in OpenType format [18].

**Synthesis**:

In the Arabic script, which is cursive, a word can be stretched by the Kashida to cover more space, and can be forced by the use of the ligatures [15]. These mechanisms can influence the sizing and the positioning of the Arabic diacritics [8].

- The justification plays an important role in the positioning of diacritics, which is not true in Latin.
- The first adaptation of the optimum fit algorithm to Arabic was made by ignoring diacritics.

## 5. Design of Diacritical Marks

The design of Latin diacritics has three challenges:
- be harmonious with the base glyph;
- collocation with other diacritics on same base glyph;
- respect the baseline and interline.

In Arabic, the design of diacritics has a supplemental challenge: it must to be harmonized with the diacritics of neighboring glyphs and ligatures, and fill the space. There are also aesthetic diacritics whose positions depend on other diacritics. The relationship between interactive diacritics and the mechanisms of multilevel and justification require resizing and repositioning of diacritics in an influenced word. Below, we present the main issues of design diacritics [4] and the specific problems to Arabic.

### 5.1 Latin case

**Asymmetry**: *Balance* is defined as an equanimity resulting from the review of an image in relation to ideas of the visible structure [19]. That is the grouping of entities in a design required on the report of their weight in a configuration of a visual picture. Balance generally is of two kinds: symmetrical and asymmetrical. The *symmetrical balance*, or *formal balance*, take place when the weight of a graphic composition is one and the same divided on every side of an invisible central axis that can be vertical or horizontal. The *asymmetrical balance*, or *informal balance*, exists when the weight of the graphic composition is not spread equally surrounding a central axis [19]. The size and weight of a Latin diacritic must be balanced with the base glyph [4]. The horizontal alignment of a diacritical glyph should be such that there is balance between the diacritic and base glyph. To symmetrically balance, a diacritic simply align the center of the bounding box of the diacritic with the base glyph [2] [4]. If either one is asymmetrical other means must be turned to account.

If the base glyph is symmetrical, *Optical alignment* is a tool to adjust the horizontal displacement of a base glyph or diacritic to focus on the diacritic glyph and maintain basic balance. Among the solutions, one is to align the optical center of the glyph with the mathematical center of space [4]. The optical center is estimated by the center of the contour (see Fig. 18).

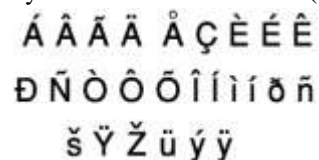
Fig. 18. Symmetrical base glyphs

If the base glyph is asymmetrical, the diacritic may connect to the following base glyph. The optical alignment is not always used and other solutions are offered by new technologies, such as OpenType.

**Harmonization**: When the diacritics are sufficiently harmonized with the corresponding base glyph, there are sometimes problems with neighboring base glyphs. For example, the tilde may touch the neighboring base glyph "U" (see Fig. 19).

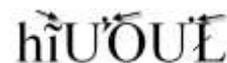
Fig. 19. Conflict of diacritics with other glyphs

One solution is to draw the diacritic specific to each base glyph, reducing the size of diacritics. Another solution is to use *kerning* [2] [4].

**Vertical space**: In some fonts, the diacritics are aligned on a line parallel to the baseline. In other fonts, the distance between the diacritic and its base glyph is variable.

**Multiple diacritics**: Multiple diacritics can cause many problems with the baseline, with other glyphs, or amongst other diacritics. Different techniques are used to solve this problem including: drawing a glyph with all the diacritics together [4].

5.2 Specific issues to Arabic case

Additionally, Arabic diacritics' role is to fill space in a word where there are specific diacritics, or they may be added for aesthetic reasons. The principal mechanisms to create space in the Arabic word are: to use the Kashida, to adjust the glyph, and to modify the interconnection between glyphs. The space can be filled by following the steps below:
 (1) Put dots and/or Shadda at the center of space.
 (2) Resize the diacritic Fatha or Fathatan proportionally with space.
 (3) Reposition the resized diacritic.
 (4) Reposition the other diacritics.
 (5) Place aesthetic and explanatory diacritics in the space.

Diacritics tend to reflect characteristics common to many glyphs, acting according to their function in a language.

The concept of symmetry in Arabic design is related to the linear writing where extensions are used to balance the masses of other glyphs.

**Word as mass**: The process of composing characters, above the line, affects the aesthetics of Arabic writing. So we have to study the space in words in relation to the sequence of their characters.

**Relationship between characters**: The relationship between characters of the word is based on (see Fig.20):
- *Identicalness*: when many of the glyphs involved in the forms, or in the forms of their parts are the same.
- *Similarity*: when letters require manual rules to be joining.
- *Harmony*: when most of the characters appear on the baseline.
- *Contrast*: when conflict is present between straight characters, horizontals and rounds characters.

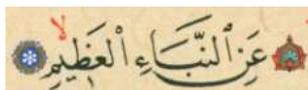

Fig. 20.   Relationship between characters

**Priority to a language's diacritics**: Diacritics lead to a repetition of common characteristics among many letters. They must not come into conflict with the diacritics of neighboring base glyphs.

**Additional role of aesthetic diacritics**: Aesthetic diacritics must be positioned to maintain symmetry and harmony in relation to a language's diacritics (see Fig. 21).

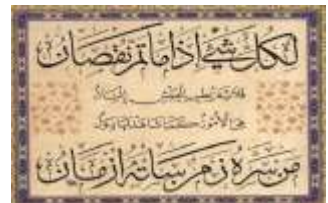

Fig. 21.   Arabic diacritics' roles

## 6. Rendering of diacritics

The factors mentioned above must be taken together in order to properly render Arabic diacritics. There are tow category of fonts: *dumb fonts* and *smart fonts*. The first category is characterized by simple sequential positioning, while the second include the complex positioning data. We will survey the various possibilities offered by the second category and their limitations in trying to represent Arabic writing properly.

6.1 Diacritics positioning processing

The non smart fonts are very limited to positioning properly the diacritics. The diacritics placement processing is designed to be used with such font format. To place one or more diacritics, this processing uses a diacritic's bounding box, the base glyph's bounding box, and a diacritic data (see Fig. 22). When the processing receives the information that the mark is to be placed over the base glyph, it looks up the orientation for this mark in tables. Based on this information, the processing calls a pair of functions, H and V, for properly positioning the mark [20].

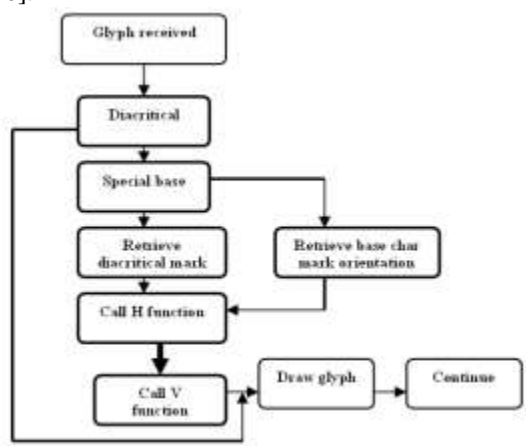

Fig. 22.   A diacritics positioning processing

To extend a processing which operates under the same architecture to be as Arabic diacritics positioning processing, the following issues must be taken into account:
- Ability to calculate the horizontal and vertical position of diacritic glyph relative to the base glyph and the neighboring base glyphs.
- Ability to calculate the horizontal and vertical position of diacritic glyph relative to the diacritics of neighboring base glyphs.

- Ability to substitute the diacritic variant if a Kashida or a ligature takes place.
- Ability to keep the contextual form of the basic glyph.

## 6.2 Rendering processing with smart font

There are many smart fonts, such as: OpenType, Graphite, and AAT, all based on Unicode. We have chosen the OpenType, as it is the most common. OpenType is a font format developed jointly by Adobe and Microsoft. It is an extension of the TrueType font format, adding support for PostScript font data. It is organized by *script*, *language system*, *feature*, and *lookup*. The notion of script denotes a collection of glyphs used to represent one or more languages in written form. A language system changes the functions or appearance of glyphs in a script to represent a given language by defining features which are typographic rules for using glyphs [21]. A feature groups the rules stocked in the font that the engine rendering execute in three phase:

**Substitution phase**: corresponds to GSUB table, which have charge to ligatures, contextual forms, vertical rotation, conversion to small caps, Indic glyphs rearrangement, etc. The principal substitutions offers are [22]:

- *Single substitution*: permits alternating from one glyph to another.
- *Multiple substitutions*: permits changing one glyph by a others.
- *Alternate substitution*: provides having a series of alternates for each glyph.
- *Ligature substitution*: permits alternating a string of glyphs with another glyph.
- *Contextual substitution*: assigns substituting a string of glyphs by another string of glyphs.

**Position phase**: corresponds to GPOS table, which manages the positioning of glyphs. We can put any diacritic on any base glyph [22]. Diacritics are distributed into various classes in conformity with their behavior. Each base glyph has attachment points in a diacritic class [21] [22]. The principal lookups offers are:

- *Single adjustment*: enables replacing the metrics of a specific character.
- *Pair adjustment*: authorize substituting the metrics of a specific pair of glyphs.
- *Cursive attachment*: permits forcing adjacent glyphs to join at specific dots.
- *Mark to base*: assigns how diacritics are placed over base glyphs.
- *Mark to ligature*: allows how diacritics are positioned over a ligature and may have various places at which the same type of mark may be positioned.
- *Mark to Mark*: provides how many diacritics are placed over base glyph.

**Justification phase**: corresponds to JSTF table, which gives the composition software means to increase or reduce the width of certain words to get the best spaces between words, in an attempt to justify a text [21].

Kashida, ligature, and allograph can be managed by GSUB and GPOS. Positioning and resizing diacritics over them can also be created by these tables.

## 6.3 Diacritics and GPOS

There are three lookups in GPOS table that threat a diacritical positioning. But before exploring them, let's see the structure of each lookup.

**Lookups Structure**: In OpenType, each lookup contains the followed elements [21]:

- LookupType: determines the type of lookup.
- LookupFlag: determines the series of flags.
- Coverage table: specifies all the glyphs are concerned by a substitution or positioning operation.

In OpenType, glyphs are divided into four types: base glyphs, diacritics; ligatures, and components of ligatures.

We can restrict the application of a lookup in some classes by Lookupflag.

**MarkToBase Attachment lookup**: is based on the followed principle: Each mark has an anchor point and associated with a class of diacritics. Every base glyph has many attachment points as there are classes of diacritical. This lookup contains a subtable MarkBasePos that composed on [21]:

- coverage table for marks;
- coverage table for base glyph;
- coordinates of the attachment points of marks;
- coordinates of the attachment points of base glyph.

**MarkToLigature Attachment Positioning Subtable**: prescribes ligatures composed of many components that can each define an attachment point for each class of marks. We find [21]:

- Coverage table for ligatures;
- Coverage table for marks;
- The attachment points for each component of each ligature.

**MarkToMark Attachment Positioning Subtable**: has same structure as the MarkToLigature Attachment Positioning Subtable, except that for marks we are tow tables coverage for marks and same for coordinates [21].

## 6.4 Synthesis

The various issues of diacritics in new technologies can be summarized in the following items:

- There is no relation of positioning diacritics (one to one) during justification;
- Diacritics and ligatures;
- Diacritics and diacritics above same base glyph.

## 7. Proposed Solution

There is no composition system which takes into account the insertion of Kashida or respecting the multilevel with position and size variation of diacritics. There is no algorithm or system to approximate space or to fill it. In this section, we present our proposed solution to approximate the void and to fill it by repositioning and resizing diacritical marks.

## 7.1 Positioning diacritics in Arabic fonts

To illustrate rendering of Arabic diacritics, we present some fonts showing their treatment of positioning diacritics.

- Traditional Arabic:

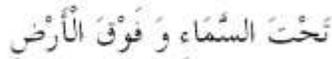

- Times new roman:

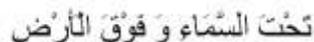

- Scheherazade:

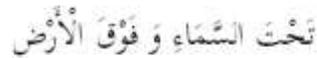

- Lateef:

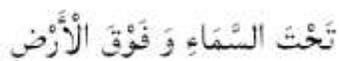

Commentary:
- Most of Arabic fonts are mono-lines.
- Some of them raise a problem during the construction of ligatures with diacritics.
- Some of them offer more than one position for diacritics.
- The different Arabic fonts do not offer a mechanism for resizing diacritics.
- The different Arabic fonts do not offer a mechanism to fill space by diacritics.

## 7.2 Font

In this section, our calligraphic proposals are based on Chawki[1] samples, *amchak* [23].

In order to determine the factors that influence specifically on the position and size of a diacritical mark, we have classified the base glyphs according to their heights and widths.

We note that:
- In isolated case, the position of the diacritical mark is associated with the width and height of the base glyph, i.e. these two factors determines the mass of the base glyph and they link with the space which corresponds to it. And with the white that precede it and/or succeed it. The size of diacritic is default if the base glyph is none stretched, and vice versa.
- In cursive forms, the position and the size of diacritics depends on the mass of its base glyph, and on those neighboring base glyphs and positions of their diacritics. The situation becomes more complicated if the multilevel takes place. We limit the study to case of single level.
- Only Fatha and Fathatan can be elongated.

We adopt that the size of Fatha and Fathatan, in depend on mass of base glyph and the mass of neighbors base glyphs.

In cursive forms, the size of the diacritic depends on the size and height of its base glyph, and on those neighboring base glyphs and positions and sizes of their diacritics [23].

**Determine diacritics position**: To take a usual distance between diacritics and base glyph requires classification of base glyphs according to their heights.

**Determine diacritics size**: There are three variants, related to its size, for Fatha and Fathatan: normal, medium and large. However, there is one for each of the others (see Fig.23).

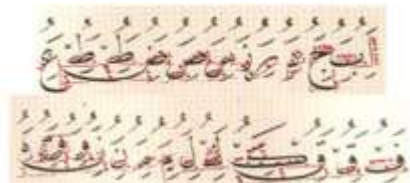

Fig. 23.   Position and size of diacritics in isolated form

Some classifications must be taken in the preparation of the font, for that we:
- Classify letters according to their mass, in each form.
- Classify letters according to their possible stretch, in each form.
- Determine, in default size, some variant positions for each diacritic, following the mass of each glyph.
- Determine, for Fatha and Fathatan, some variant sizes, following each stretched letter class.
- Determine the pairs of glyphs where there resizing Fatha.
- Determine the positions of the diacritics of these pairs and the possible alternatives.

## 7.3 Algorithm

The proposed algorithm aims to provide a mechanism to position Arabic diacritics with the proper size to fill space that is influenced by the effects of the justification and the ligature. In this algorithm, we have adopted two principles: First, change the positions and dimensions of diacritics, related to the mass of the base glyph, to the mass of the followed glyph, and to the difference of those heights. Second, consider ligatures as a result of a series of basic glyphs which each has a diacritic of its own [24].

The algorithm:
- In the first phase:
```
(1) Put the suitable position diacritic, with
    the default size, for the first glyph of
    word.
```
- In the phase n, where n ≥ 2 as long as writing the word has not been completed:
```
(1) Put the suitable position diacritic, with
    the default size, for the current glyph.
(2) If the diacritic of the preceding glyph
    is Fatha or Fathatan, reposition and
    resize it and reposition the diacritic of
    current glyph according to fill space,
```

---
[1] Mohamed Chawki (1828-1887), a great calligrapher Turk and famous in the history of Arabic calligraphy, was certified in calligraphy at the age of 13 years.

```
        else reposition the diacritic according
        to fill space.
    - In the final phase:
    (1) If the diacritic of the last character is
        Fatha or Fathatan, call the alternative
        diacritic related to its mass.
```

The chose of suitable, position and size, diacritic will be a processing of substitution in tables offered by font.

Following the text graphic context (illustration, handbook, lecture note, book, masterpiece…), the user would choose between activate/deactivate calligraphic treatments (ligature, Kashida, allograph, etc.). However, the user, or writer, needs to be qualified in order to decide the suitable convention. This chosen convention will influence the strategy adopted in the justification processing. In calligraphy, there is no priority strategy. The publishing system should be able to invoke, or re-invoke, the positioning and sizing of diacritics, after running the composition and justification processing.

7.4 Results

The algorithm has been applied on an Arabic font, developed in OpenType format, and we have a result example shown in Fig. 24. The obtained results indicate that OpenType is limited by:

- Substitution: The using of the Kashida by substitution glyph, to extend, by an extended variant, must be done after a good choice of location in where to put it. The Kashida was adapted in [18] considering it to be part of the extended glyph.
- Reorganization: There was only one of two options:
  o *Ignore the ligatures*: Then, two cases have in place: not aware of multilevel, which gives incomplete results; or split the ligatures to glyphs, but it affects the rendering engine.
  o *Takes the ligature into account*: However, with reorganization and a change in the order of diacritics and base glyphs to form the ligatures with their diacritics.

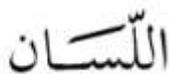

Fig. 24.    Resizing Fatha

## 8. Conclusion

In this paper, the problem of elongation of diacritical marks is treated by simplifying the contexts of elongation of the marks in a context of neighborhood in two successive letters. But this is not the case; there are diverse contexts which control the choice of the size of a diacritical mark, also a context of neighborhood contains often more than two successive letters. The proposed algorithm allows stretching out the size of a diacritic, but without offering a mechanism neither to insert aesthetic objects nor to offer a tool to straighten the positions of the nearby diacritical marks. The problem must be studied in a more systematic approach by basing itself on a study concerning the choice of the sizes of diacritical marks and their positions. The positioning and resizing of Arabic diacritics is related to the effects of writing in cursive, multilevel ligature and justification by the Kashida. These factors depend on Arabic calligraphic styles; each one is controlled by its own rules. Consequently, the Arabic script must be treated as a set of styles in the electronic publishing systems.

The text composition systems underspecified the complicated positioning of diacritics when compared to ligation. Diacritics positioning can be lost when diacritics are repositioned over glyphs in cursive attachment. Remember that attachment is a cursive smart font feature that allows the attachment and positioning of glyphs and determines how to find the connecting dot of the neighboring glyphs; it is not just a simple alignment of letters on the baseline. Another level of complication is when a paragraph is justified. A font cannot predict the mechanism implemented by the composition engine to justify the lines, for example. The part of the engine that handles paragraphs may create elongated glyphs by Kashida, cause substitution of alternate glyphs, or permit activations/deactivations of ligatures. Each part of the system must contribute in its own way to the final visual rendering, and we must know, above all, the tasks involved in each component.

**Mohamed Hssini** is a Ph.D. student in Department of Computer Science at Cadi Ayyad University. He is a member of *multilingual scientific e-document processing* team. His current main research interest is multilingual typography, especially the publishing of Arabic e-documents while observing the strict rules of Arabic calligraphy. Email:m.hssini@ucam.ac.ma.

**Azzeddine Lazrek** is full Professor in *Computer Science* at Cadi Ayyad University in Marrakesh. He holds a *PhD* degree in *Computer Science* from Lorraine Polytechnic National Institute in France since 1988, in addition to a *State Doctorate* Morocco since 2002. Prof. Lazrek works on *Communication multilingual multimedia e-documents in the digital* area. His areas of interest include: multimedia information processing and its applications, particularly, electronic publishing, digital typography, Arabic processing and the history of science. He is in charge of the research team *Information Systems and Communication Networks* and the research group *Multilingual scientific e-document processing*. He is an *Invited Expert* at W3C. He leads a *multilingual e-document composition* project with other international organizations.
Email:lazrek@ucam.ac.ma.